\def\year{2022}\relax
\documentclass[letterpaper]{article} 
\usepackage{aaai21}  
\usepackage{times}  
\usepackage{helvet} 
\usepackage{courier}  
\usepackage[hyphens]{url}  
\usepackage{graphicx} 
\usepackage{amsmath}
\usepackage{amssymb}
\usepackage{algorithm}
\usepackage{algorithmic}
\usepackage{amssymb}
\usepackage{amsfonts}
\usepackage{amsthm,amsmath,amssymb}
\usepackage{booktabs}
\usepackage{bm}
\usepackage{caption}
\usepackage{color}
\usepackage{epsfig}
\usepackage{graphicx}
\usepackage{multirow}
\usepackage{multicol}
\usepackage{mathrsfs}
\usepackage{pifont}
\usepackage{subfigure} 
\usepackage{textcomp}
\usepackage{threeparttable}
\usepackage{url}
\usepackage{xcolor}
\usepackage[colorlinks,linkcolor=black,anchorcolor=black,citecolor=black]{hyperref}
\usepackage[switch]{lineno}

\definecolor{qiu}{RGB}{250,10,10}

\definecolor{fan}{RGB}{181,68,52}

\newcommand{\cmark}{\ding{51}}%
\newcommand{\xmark}{\ding{55}}%

\usepackage{natbib}  
\usepackage{caption} 
\title{Generation, augmentation, and alignment: A pseudo-source domain based method for source-free domain adaptation \footnote{Under review}}
\author{
    Yuntao Du, 
    Haiyang Yang, 
    Mingcai Chen,
    Juan Jiang,
    Hongtao Luo,
    Chongjun Wang
    \\
}
\affiliations{

    Department of computer science and technology, Nanjing University\\
    \{duyuntao,mf20330105,chenmc,MF20330037,mf20330054\}@smail.nju.edu.cn, chjwang@nju.edu.cn

}

\begin{document}
\maketitle

\begin{abstract}
%
Conventional unsupervised domain adaptation (UDA) methods need to access both labeled source samples and unlabeled target samples simultaneously to train the model. While in some scenarios, the source samples are not available for the target domain due to data privacy and safety. To overcome this challenge, recently, source-free domain adaptation (SFDA) has attracted the attention of researchers, where both  a trained source model and unlabeled target samples are given. Existing SFDA methods either adopt a pseudo-label based strategy or generate more samples. However, these methods do not explicitly reduce the distribution shift across domains, which is the key to a good adaptation. Although there are no source samples available, fortunately, we find that some target samples are very similar to the source domain and can be used to approximate the source domain. This approximated domain is denoted as the pseudo-source domain. In this paper, inspired by this observation, we propose a novel method based on the pseudo-source domain. The proposed method firstly generates and augments the pseudo-source domain, and then employs distribution alignment with four novel losses based on pseudo-label based strategy. Among them, a domain adversarial loss is  introduced between the pseudo-source domain the remaining target domain to reduce the distribution shift. The results on three real-world datasets verify the effectiveness of the proposed method. 
 \end{abstract}

\section{Introduction}

Deep neural networks
need large labeled samples to train the model \cite{He2016DeepRL,Ren2015FasterRT} and the performance will drop seriously due to the domain shift when applying the trained model directly to a new domain. As a promising learning paradigm, unsupervised domain adaptation (UDA), a sub-field of transfer learning, is able to transfer knowledge from the labeled source domain to the unlabeled target domain to overcome this challenge \cite{Pan2010ASO}.

Conventional deep unsupervised domain adaptation methods are built on two main strategies: moment matching \cite{Long2015LearningTF, Kang2019ContrastiveAN,Chen2020HoMMHM} and adversarial domain adaptation \cite{DBLP:journals/jmlr/GaninUAGLLML16,saito2018maximum,zhang2019bridging}. The former minimizes the statistical distribution discrepancy across domains and the latter reduces the domain discrepancy in an adversarial manner. And
existing UDA methods need to access both the source samples and target samples simultaneously to train the model. However, in some scenarios, the source model instead of the source samples is able to be obtained due to data privacy and safety.  Such setting is denoted as source-free domain adaptation (SFDA) \cite{liang2020shot,Li2020ModelAU}. Previous UDA methods cannot be applied to SFDA directly due to the absence of source samples.

Generally,  the goal of SFDA is to train a well-performed model in the target domain based on a trained source model and unlabeled target samples. There are some attempts to tackle this problem. 
Both SHOT \cite{liang2020shot} and CPGA \cite{Qiu2021SourcefreeDA} adopt pseudo-label based strategy together with information maximization and prototypes adaptation respectively.  
Besides, both MA \cite{Li2020ModelAU} and SDDA \cite{Kurmi2021DomainIA} adopt a generative adversarial net to generate either target-like or source-like samples to train the model. 
However, these methods do not explicitly reduce the distribution shift across domains. According to classical UDA theories \cite{BenDavid2009ATO,zhang2019bridging}, it is important to reduce the distribution discrepancy across domains to achieve a good adaptation and such property should also be meet in SFDA. 
So the large domain shift is a limitation for existing SFDA methods and needs to be addressed.

\begin{figure*}[t]
	\centering
	\includegraphics[width=0.9\textwidth]{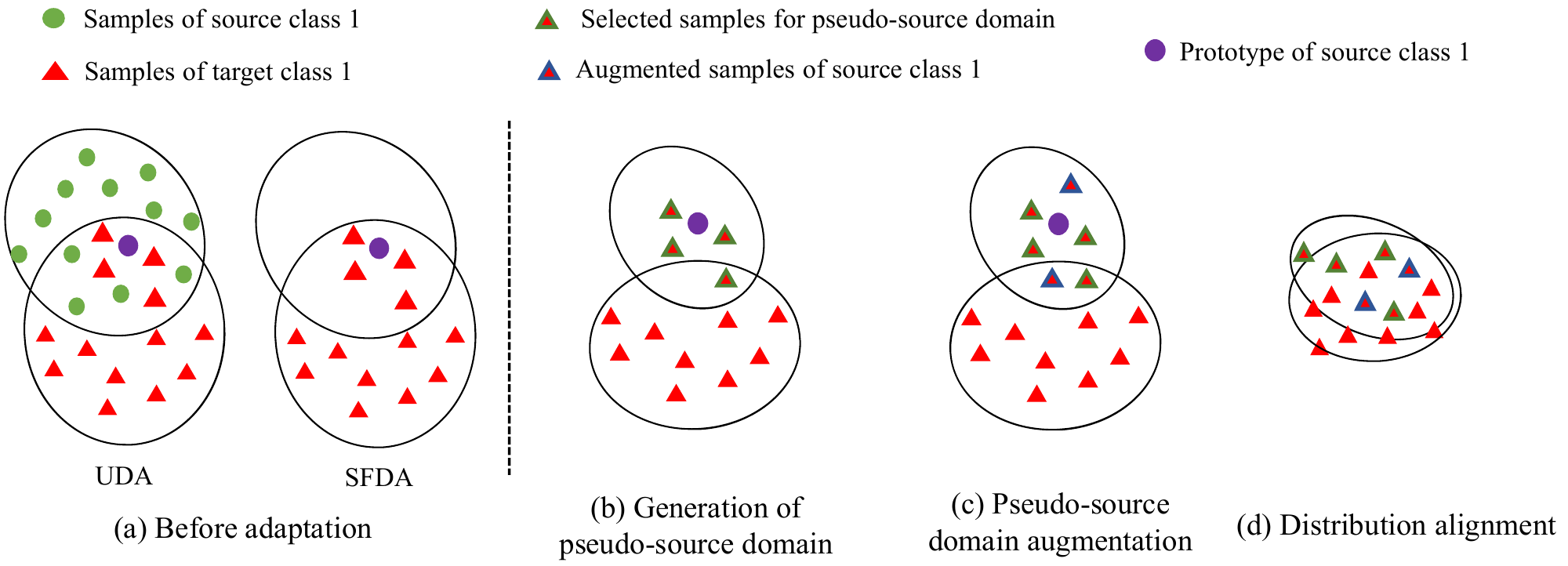}
	\vspace{-1mm}
	\caption{The motivation of the proposed method. (a) Although there are no source samples, we find that the target samples near the source prototypes can be used to represent the source domain. Thus, (b) PS firstly generates a pseudo-source domain by selecting such target samples. Then, (c) PS enlarges this pseudo-source domain using data augmentation. Lastly, (d) PS performs distribution alignment with four losses. Among them, the domain adversarial loss is introduced to reduce the distribution shift.}
	\label{fig:exam}
	\vspace{-4mm}
\end{figure*}

Based on the above analysis, we aim to reduce domain shift  to further improve the performance of SFDA.  Although there are no source samples, the trained model can retain and reflect original data distribution information. As shown in Figure \ref{fig:exam}(a), assuming there are some source samples, it is obvious that they are spread around the corresponding prototypes, where the class centers or weight vectors of the classifier can be regarded as the prototypes \cite{saito2018maximum}. 
Although there exists domain shift, we obverse that some target samples are also spread around the corresponding source prototypes and they are very similar to the source domain. Thus, these target samples could be used to approximate the source domain. We denote these samples as pseudo-source samples and the corresponding new domain as the pseudo-source domain.
In other words, the target samples could be split into two disjoint parts, i.e., the pseudo-source samples and the remaining target samples.
In such cases, reducing the distribution discrepancy across domains can be approximately achieved by minimizing the discrepancy between the pseudo-source samples and the remaining target samples. Although some UDA methods also split the target samples into two parts, such as easy samples and hard samples in \cite{Pan2020UnsupervisedIA}, the goal is different. The method in \cite{Pan2020UnsupervisedIA} aims to reduce the intra-domain gap within the target domain while our method aims to generate pseudo-source samples for reducing the shift across domains.

Following the above idea, in this paper, we propose a method named \emph{Pseudo-Source domain based source-free domain adaptation} (PS). 
Our method is composed of three steps. The first step is to generate a pseudo-source domain by selecting the target samples that are similar to the source domain. As the number of the samples in the pseudo-source domain is smaller than the number of the remaining target samples, the second step is to augment the pseudo-source domain to enlarge this new domain for better adaptation. The third step is to perform distribution alignment with four proposed losses.
During distribution alignment, PS also adopts pseudo-label based strategy to train the model. Notably, among four losses, a domain adversarial loss is introduced between the pseudo-source domain and the remaining target domain to decrease the distribution discrepancy across domains. To sum up, our principal contributions are summarized as follows:
\begin{itemize}
    \item  We are the first to generate a pseudo-source domain from the target samples and perform distribution alignment for SFDA, such that the domain discrepancy across domains can be reduced even without source samples.
    \item  We propose a novel method, which generates and augments the pseudo-source domain and then performs distribution alignment for explicit adaptation.
    \item Extensive experiments are conducted and the results show that the proposed method achieves better performance than state-of-the-art methods.
\end{itemize}

\section{Related Work}

\subsection{Unsupervised Domain Adaptation} 

A classical domain adaptation theory \cite{BenDavid2009ATO} indicates that it is crucial to reduce the distribution discrepancy across domains to achieve better adaptation. Based on this theory, many domain adaptation methods have been proposed and they are divided into moment matching and adversarial domain adaptation. The goal of the former is to reduce the statistical distribution discrepancy across domains. The widely used statistical measurements include  the first-order moment \cite{Long2015LearningTF, Kang2019ContrastiveAN}, the second-order moment\cite{Sun2016ReturnOF}, high-order moment\cite{Chen2020HoMMHM,CQQQ} and other statistical measurements \cite{Long2017DeepTL,Li2020MaximumDD,Shen2018WassersteinDG}.  

Adversarial domain adaptation, which is inspired by generative adversarial network \cite{goodfellow2014generative}, reduces the distribution discrepancy in an adversarial manner. DANN \cite{DBLP:journals/jmlr/GaninUAGLLML16} introduces a domain discriminator which plays a min-max game with the feature extractor by the domain adversarial loss. MCD \cite{saito2018maximum} introduces two classifiers as a discriminator to play a min-max game with the feature extractor such that the source samples are pushed into the support of the target domain. Considering practical multi-class problem, MDD \cite{zhang2019bridging} proposes a margin-based theory, and a new method based on this theory is proposed. Following methods \cite{Cicek2019UnsupervisedDA,Chen2020UnsupervisedDA} adopt the multi-class discriminator which considers the domain and class information simultaneously. However, previous UDA methods cannot be applied to SFDA directly due to the absence of source samples.

\subsection{Source-Free Domain Adaptation}
The source samples are unavailable in source-free domain adaptation due to data privacy and security.  However, a model trained by the source samples is given and the goal is to adapt the trained model to the target domain. Some methods adopt pseudo-label based  strategy. SHOT \cite{liang2020shot}  learns a target-specific feature extraction module by self-supervised pseudo-labeling together with information maximization and implicitly aligns representations of the target domain to the source model.  CPGA \cite{Qiu2021SourcefreeDA}  proposes to utilize the hidden knowledge in the source model and exploits it to generate source avatar prototypes as well as target pseudo-labels for domain alignment.  PrDA \cite{kim2020progressive} progressively updates the target model in a self-learning manner by leveraging a pre-trained model from the source domain, such that the pseudo-labels would become more accurate for better adaptation.

Some methods adopt generative adversarial nets (GAN) to generate more samples.
MA \cite{Li2020ModelAU} proposes to generate target-style data with a class conditional generative adversarial net.  SDDA \cite{Kurmi2021DomainIA} treats the trained classifier as an energy-based model to learn the data distribution along with a generative adversarial net for generating source-like samples. Besides, some works explore source-free domain adaptation under different settings such as universal SFDA \cite{Kundu2020UniversalSD} and multi-source SFDA \cite{Ahmed2021UnsupervisedMD,Feng2021KD3AUM}. Moreover, Some works focus on different applications such as  semantic segmentation \cite{Liu2021SourceFreeDA,Zhao2021SourceFreeOC} and object detection \cite{Li2021AFL}. Although achieving remarkable progress, these methods do not explicitly reduce the distribution shift across domains.


\begin{figure*}[t]
	\centering
	\includegraphics[width=0.72\textwidth]{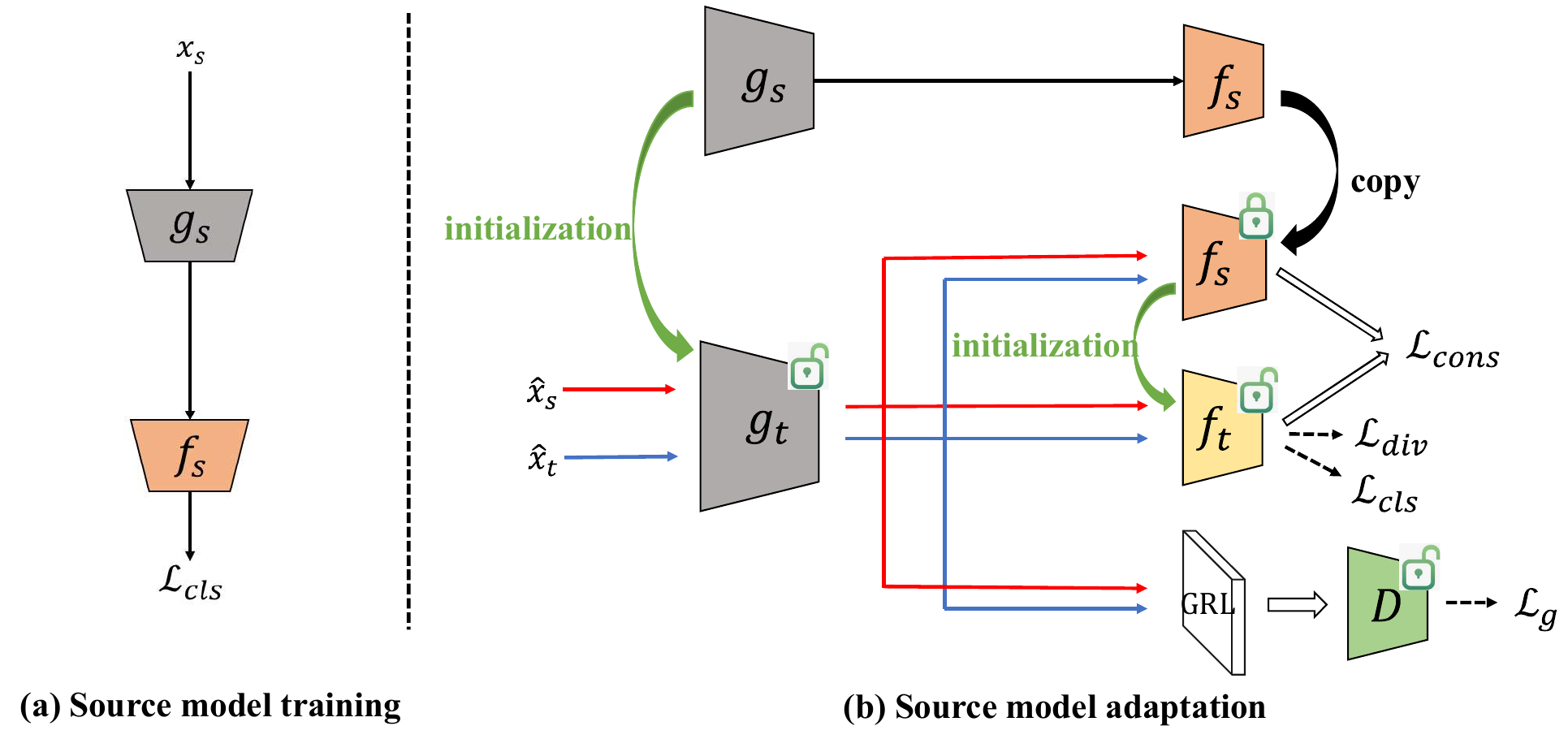}
	\vspace{-1mm}
	\caption{The model structure of the proposed method.  $g_s$, $f_s$ and $g_t$, $f_t$ are the feature extractors and classifiers of the source model and target model, respectively. GRL is the gradient reversal layer and $D$ is the domain discriminator.}
	\label{fig:model}
	\vspace{-5mm}
\end{figure*}


\vspace{-2mm}
\section{Problem Definition}

In a vanilla UDA task, there are a labeled source domain $\mathcal{D}_s = \{x_s^i, y_s^i\}_{i=1}^{n_s}$ and an unlabeled target domain $\mathcal{D}_t = \{x_t^i\}_{i=1}^{n_t}$, where $x_s^i \in \mathcal{X}_s$, $y_s^i \in \mathcal{Y}_s$, and $x_t^i \in \mathcal{X}_t$. The feature space and the label space are the same across domains, i.e.,  $\mathcal{X}_s = \mathcal{X}_t = \mathcal{R}^d$ and $\mathcal{Y}_s = \mathcal{Y}_t = \{1,2,...,K\}$. While the source samples and target samples are drawn from different distributions $P_s(x,y)$ and $P_t(x,y)$, i.e., $P_s(x,y) \neq P_t(x,y)$. In UDA, the goal is to train a model that can perform well in the target domain using both labeled source samples of $\mathcal{D}_s$ and unlabeled target samples of $\mathcal{D}_t$. 

While in SFDA, the target domain can not access the source samples, but a model $F_s$ trained by the source samples is given. The goal of SFDA is to use the trained source model $F_s$ and unlabeled target samples of $\mathcal{D}_t$ to train a new model $F_t$, such that $F_t$ performs well in the target domain.

\vspace{-2mm}
\section{Method}

The model structure of the proposed method is shown in Figure \ref{fig:model}(b). The source trained model $F_s$ is composed of  a  feature  extractor $g_s$ and  a  classifier $f_s$, i.e., $F_s(x) =f_s(g_s(x))$. 
The target model also contains a feature extractor $g_t$ and a classifier $f_{t}$, i.e., $F_t(x) =f_t(g_t(x))$. Besides, a domain discriminator $D$ is also introduced. 
To better transfer knowledge from the source domain to the target domain, the feature extractor $g_t$ of target model  is initialized by the feature extractor $g_s$ of source model, and the target classifier $f_{t}$ is also initialized by the source classifier $f_s$. Note that the source classifier $f_{s}$ is fixed during training, while the target classifier $f_{t}$ would be updated. In the following subsections, we firstly introduce the training process of source model and then describe these training steps of the proposed method.

\subsection{Training of Source Model}\label{source_model_g}
In SFDA setting, as is shown in Figure \ref{fig:model}(a),  a model $F_s: \mathcal{X}_s \rightarrow \mathcal{Y}_s$ is trained by the source samples and  the trained model $F_s$ is delivered to the target domain. 
%
%
To further increase the discriminability of the source model, following previous work \cite{liang2020shot}, PS adopts the label-smoothing technique to train the model as it encourages samples to lie in tight
evenly separated clusters \cite{Mller2019WhenDL}. The objective function is,
\begin{equation}
		\begin{aligned}
			\mathcal{L}_{cls}^{s}(g_s, f_s; \mathcal{D}_s) =& \\
			-\mathbb{E}_{(x_s,y_s)\in \mathcal{D}_s }& \sum\nolimits_{k=1}^{K} q_k^{ls} \log \delta_k(f_s(g_s(x_s)))
		\end{aligned}
	\end{equation}  
where $f_s(g_s(x_s))$ is the $K$-dimensional output of the source sample $x_s$.  $q^{ls}_k=(1-\gamma)q_k^s + \gamma/K$ is the smoothed label and $\gamma$ is the smoothing parameter which is empirically set to 0.1. $\delta_{k}(\mathbf{a}) = \frac{
exp(a_k)} {\sum_i exp(a_i)} $ denotes the $k$-th element in the softmax output of a $K$-dimensional vector $\mathbf{a}$.  $q^s$ is the one-of-$K$ encoding of $y_s$ where $q_k^s$ is ‘1’ for the correct class and ‘0’ for the rest. 

\subsection{Generation of Pseudo-Source Domain}\label{gene_source}

As the source samples are not available in the target domain,  conventional UDA methods can not be directly used to reduce the domain shift. Fortunately, we find that some target samples can be used to represent the source domain. Such samples are  denoted as pseudo-source samples and constitute a new domain denoted as pseudo-source domain.
This process can be regarded as dividing the samples in the target domain into two disjoint parts, where the first part is the pseudo-source domain, denoted as $\hat{\mathcal{D}}_s$, 
%
%
and the second part is the remaining target domain, denoted as $\hat{\mathcal{D}}_t$. It is obvious that $\hat{\mathcal{D}}_s \cap \hat{\mathcal{D}}_t = \varnothing$ and $\hat{\mathcal{D}}_s \cup \hat{\mathcal{D}}_t =\mathcal{D}_t$.

As is shown in Figure \ref{fig:exam}, the target samples near the source prototypes are  similar to the source domain. The entropy criterion is widely used in  UDA \cite{long2018conditional,Pan2020UnsupervisedIA} to evaluate the uncertainty of classifier prediction for a given sample. The lower the entropy is, the closer the sample is to the prototype. Thus, in this paper, PS adopts the entropy criterion to select the target samples that can be used to simulate the source domain. Given a  trained source model $F_s$, all the target samples firstly pass the source model and then the entropy $H(x_t)$ is calculated, 
\begin{equation}
    H(x_t) = -\sum\nolimits_{k=1}^K \delta_{k}(f_s(g_s(x_t)))\log\delta_{k}(f_s(g_s(x_t)))
    \label{entro}
\end{equation}
%

 To generate a class-balanced pseudo-source domain, 
 all the target samples are sorted in ascending order of entropy and
 PS selects samples from each class separately, i.e., the top  $\alpha$ proportion of samples in each class are selected. As the samples in the target domain are unlabeled, the pseudo-labels are used instead. For the sample ${x}_t \in \mathcal{D}_t$, the pseudo-label $\hat{y}_t$ is obtained by 
\begin{equation}
    \hat{y}_t = \arg\max_{y \in \{1,...,K\}}f_s(g_s(x_t))
    \label{pseudo_1}
\end{equation}
By this strategy, PS generates a class-balanced new domain and we experimentally find that $\alpha = 0.1$ works well.


\subsection{Pseudo-Source Domain Augmentation}
Previous works have shown that the larger the source domain is, the better performance the model could get \cite{DBLP:journals/jmlr/GaninUAGLLML16,Kumar2018CoregularizedAF}. However, the number of the samples in the pseudo-source domain $\hat{\mathcal{D}}_s$ is smaller than that in the remaining target domain $\hat{\mathcal{D}}_t$ (namely, $|\hat{\mathcal{D}}_s|:|\hat{\mathcal{D}}_t|\approx$ 1:9). Thus, PS enlarges the pseudo-source domain using data augmentation. In this work,  mixup \cite{Zhang2018mixupBE} is choosen, as it is a simple yet effective supervised data augmentation method. 
Given two samples from the pseudo-source domain, e.g., $\hat{x}_s^i, \hat{x}_s^j\in \hat{\mathcal{D}}_s$, whose pseudo-labels are obtained by Equ \ref{pseudo_1} and denoted as $\hat{y}_s^i$ and $\hat{y}_s^j$, PS augments the pseudo-source domain by  mixing these two samples,
\begin{equation}
    \begin{split}
    & \hat{x}_{s,aug} = \lambda \hat{x}_s^i + (1 - \lambda) \hat{x}_s^j \\
    & \hat{y}_{s,aug} = \lambda \hat{y}_s^i + (1 - \lambda) \hat{y}_s^j 
    \end{split}
    \label{mix}
\end{equation}
where $\lambda \sim $ Beta($\beta$, $\beta$), for $\beta \in (0, \infty)$. 
The new domain which is composed of augmented pseudo-source samples is denoted as augmented domain $\hat{\mathcal{D}}_{aug} = \{\hat{x}_{s,aug},\hat{y}_{s,aug}\}_{i=1}^n$. And the domain that contains the original pseudo-source domain $\hat{\mathcal{D}}_s$ and the augmented source domain $\hat{\mathcal{D}}_{aug}$ is denoted as augmented pseudo-source domain $\hat{\mathcal{D}}_{s}^{aug}$, i.e. $\hat{\mathcal{D}}_{s}^{aug} = \hat{\mathcal{D}}_s \cup \hat{\mathcal{D}}_{aug}$. After adopting mixup,  the dataset can be effectively enlarged.

\subsection{Distribution Alignment}\label{da}
%
After generating and augmenting the pseudo-source domain, PS performs distribution alignment where pseudo-label based strategy is also used to train the model. In distribution alignment, four complementary  losses are proposed for better adaptation in the target domain.


\subsubsection{Classification loss}
PS firstly trains the new model $F_t(x)$ to classify the samples in both the augmented pseudo-source domain  and the remaining target domain correctly. Similar to previous work \cite{Saito2017AsymmetricTF}, training the model to classify the target samples with pseudo-labels could help the model capture the characteristics of the target domain. The objective function is
\begin{equation}
		\begin{aligned}
			\mathcal{L}_{cls}(g_t, f_t; \hat{\mathcal{D}}_s^{aug}, \hat{\mathcal{D}}_t) & = 
			\mathbb{E}_{(\hat{x}_s,\hat{y}_s)\sim \hat{\mathcal{D}}_s^{aug}} L(g_t(f_{t}(\hat{x}_s)), \hat{y}_s)  \\ + \mathbb{E}_{(\hat{x}_t,
			\hat{y}_t)  \sim \hat{\mathcal{D}}_t} & L(f_t(g_{t}(\hat{x}_t)),\hat{y}_t)
		\end{aligned}
		\vspace{-1mm}
	\end{equation} 
where $L(\cdot, \cdot)$ is the cross-entropy loss. For the samples in $\hat{\mathcal{D}}_t$, the pseudo-labels obtained by Equ \ref{pseudo_1} may be incorrect as the entropy is very high. We incorporate the method in SHOT \cite{liang2020shot} to regenerate the pseudo-labels  and the detail steps are shown in the supplementary materials.

\subsubsection{Diversity loss}
As there are only unlabeled samples in the remaining target domain, to avoid the collapse mode, PS adopts the information maximization \cite{Gomes2010DiscriminativeCB} to make the target outputs globally diverse,
\begin{small}
\begin{equation}
		\begin{aligned}
			\mathcal{L}_{div}(g_t, f_{t};\hat{\mathcal{D}}_t) &= \sum\nolimits_{k=1}^{K}p_{1,k}log p_{1,k} + \sum\nolimits_{k=1}^{K}p_{2,k}log p_{2,k} \\
			= [D_{KL} & (p_1, \frac{1}{K}\mathbf{1}_K) + D_{KL}(p_2, \frac{1}{K}\mathbf{1}_K)]-2logK,
		\end{aligned}
		\label{eq:ent}
	\end{equation}
\end{small}
where $\mathbf{1}_K$ is a $K$-dimensional vector with all ones, and $p_1 = \mathbb{E}_{\hat{x}_t \in \hat{\mathcal{D}}_t}[\delta(f_{s}(g_t(\hat{x}_t)))]$  and $p_2 = \mathbb{E}_{\hat{x}_t \in \hat{\mathcal{D}}_t}[\delta(f_{t}(g_t(\hat{x}_t)))]$ are the mean output embeddings of the whole target domain by classifiers $f_{s}$ and $f_{t}$, respectively. $p_{1,k}$ and $p_{2,k}$ are the $k$-th element of the outputs of $f_{s}$ and $f_{t}$, respectively.
With the fair diversity-promoting objective $\mathcal{L}_{div}$, the learned model can circumvent the trivial solution where all unlabeled data have the same one-hot encoding.

\begin{algorithm}[t]
		\caption{Training of PS}
		\label{algorithm_ps}
		\begin{algorithmic}[1]
			\REQUIRE Trained source model $F_s$ composed of $g_s$ and $f_s$, unlabeled target samples of $\mathcal{D}_t$, batchsize $B$, and  proportion $\alpha$.
			\ENSURE Trained target model $F_t$ composed of $g_t$ and $f_{t}$.
			\STATE \textbf{Initialization:} Initialize $g_t$ with $g_s$, initialize $f_{t}$ with $f_s$.
			\vspace{-4mm}
			\REPEAT 
			\STATE Sample a minibatch of size $B$ from $\mathcal{D}_t$.
			\STATE \textbf{Generation}: Sort all the target samples with the entropy calculated by Equ \ref{entro} in each class, and split samples of the top $\alpha$ proportion  as $\hat{\mathcal{D}}_s$ and the remaining samples as $\hat{\mathcal{D}}_t$.
			\STATE \textbf{Augmentation}: Augment the samples in pseudo-source domains $\hat{\mathcal{D}}_s$ by Equ \ref{mix} to get augmented pseudo-source domain
			$\hat{\mathcal{D}}_s^{aug}$.
 			\STATE \textbf{Alignment:} 
 			\STATE \quad 1) Train the feature extractor $g_t$ and domain discriminator $D$ by Equ \ref{adv_1}.
 			\STATE  \quad 2) Train the feature extractor $g_t$ and the classifier $f_{t}$ by Equ \ref{adv_2}.
 			\UNTIL{Maximum iteration or loss convergence }
		\end{algorithmic}
	\end{algorithm}

\subsubsection{Constrain loss}
the source classifier $f_{s}$ is kept unchanged during  training. As the source classifier $f_{s}$ could  retain the learned knowledge from the source domain, PS aims to force the learned target classifier $f_{t}$ to be not far away from the source classifier $f_{s}$. Such strategy is also used in continuous learning  and incremental learning \cite{Li2018LearningWF,Parisi2019ContinualLL}. Specially, PS trains the features extractor $g_t$ to make the outputs of these two classifiers similar,
\begin{equation}\label{eq:be}
\begin{split}
 \mathcal{L}_{\mathrm{cons}}(g_t;\hat{\mathcal{D}}_t)  = & \mathbb{E}_{\hat{x}_t \sim \hat{\mathcal{D}}_t} [ L(f_s(g_t(\hat{x}_t)),f_t(g_t(\hat{x}_t))) \\
& + L(f_t(g_t(\hat{x_t})), f_s(g_t(\hat{x}_t))) ]
\end{split}
\end{equation}

\subsubsection{Domain adversarial loss}
Though achieving remarkable progress \cite{liang2020shot,Li2020ModelAU}, previous SFDA methods do not explicitly reduce the distribution shift across domains. So in this work, PS aims to learn  domain-invariant features in an adversarial manner. Similar to previous work \cite{ganin2015unsupervised}, a domain discriminator $D$ is introduced to distinguish the augmented pseudo-source domain $\hat{\mathcal{D}}_s^{aug}$ from the remaining target domain $\hat{\mathcal{D}}_{t}$. While the feature extractor $g_t$ is trained to confuse the domain discriminator $D$. 
The adversarial objective function is,
\begin{equation}
    \begin{aligned}
	\mathcal{L}_{g}(g_t, D ; \hat{\mathcal{D}}_s^{aug}, \hat{\mathcal{D}}_{t}) &= \mathbb{E}_{\hat{x}_s\sim \hat{\mathcal{D}}_s^{aug}}
	[\log D(g_t(\hat{x}_s))] \\
	&+ \mathbb{E}_{\hat{x}_t\sim \hat{\mathcal{D}}_t}
	[\log(1 - D(g_t(\hat{x}_t)))
	]
	\end{aligned}
	\vspace{-3mm}
\end{equation}

\subsection{Training Steps}
In this subsection, we summarize the training steps of PS.  Combining  the  above  objectives  discussed  together, the overall objective is:
\begin{equation}
    \min_{g_t} \max_{D} \mathcal{L}_{cons}(g_t)  + \lambda_g \mathcal{L}_{g}(g_t,D) 
    \label{adv_1}
    \vspace{-2mm}
\end{equation}
\begin{equation}
    \min_{g_t, f_{t}}  \mathcal{L}_{div}(g_t, f_{t}) + \lambda_{c}\mathcal {L}_{cls}(g_t,f_{t})
    \label{adv_2}
\end{equation}

Following the previous method \cite{ganin2015unsupervised}, the min-max training procedure in Equ \ref{adv_1} is accomplished by applying a Gradient Reversal Layer (GRL). GRL behaves as the identity function during the forward propagation and inverts the gradient sign during the backward propagation, hence driving the parameters to maximize the output loss. During the distribution alignment, the model is trained by the losses in Equ \ref{adv_1} and Equ \ref{adv_2} alternatively.

\textbf{Remark}: In our method, PS performs pseudo-source domain generation, pseudo-source domain augmentation and distribution alignment in an alternative manner. All three steps are employed in a minibatch of size $B$. The pseudo-source code is shown in Algorithm \ref{algorithm_ps}.

\begin{table}[tbp]
\setlength{\tabcolsep}{3.0pt}
	\centering
	\small
	\scalebox{0.8}{
		\begin{tabular}{lccccc}
			\toprule
			Method (Source$\to$Target) & Source-free & S$\to$M & U$\to$M & M$\to$U & Avg.\\
			\midrule
			Source-only & \xmark  & 67.1$\pm$0.6 & 69.6$\pm$3.8 & 82.2$\pm$0.8 & 73.0 \\
			ADDA & \xmark & 76.0$\pm$1.8 & 90.1$\pm$0.8 & 89.4$\pm$0.2 & 85.2 \\
			ADR & \xmark & 95.0$\pm$1.9 & 93.1$\pm$1.3 & 93.2$\pm$2.5 & 93.8 \\
			CDAN+E & \xmark & \multicolumn{1}{l}{89.2} & \multicolumn{1}{l}{98.0} & \multicolumn{1}{l}{95.6} & 94.3 \\
			CyCADA & \xmark & 90.4$\pm$0.4 & 96.5$\pm$0.1 & 95.6$\pm$0.4 & 94.2 \\
			rRevGrad+CAT & \xmark & 98.8$\pm$0.0 & 96.0$\pm$0.9 & 94.0$\pm$0.7 & 96.3 \\
			SWD & \xmark & 98.9$\pm$0.1 & 97.1$\pm$0.1 & 98.1$\pm$0.1 & 98.0 \\
			\midrule
            SHOT-IM & \cmark & 89.6$\pm$5.0 & 96.8$\pm$0.4 & 91.9$\pm$0.4 & 92.8 \\
            SHOT    & \cmark & 98.9$\pm$0.0 & 98.4$\pm$0.6 & 98.0$\pm$0.2 & 98.4 \\
			\textbf{PS (ours)} & \cmark & \textbf{99.0}$\pm$0.1 & \textbf{98.6}$\pm$0.0 & \textbf{98.2}$\pm$0.2 & \textbf{98.6} \\
			\midrule
			Target-supervised (Oracle) & &99.4$\pm$0.1 & 99.4$\pm$0.1 & 98.0$\pm$0.1 & 98.8 \\
			\bottomrule
		\end{tabular}
		}
		\caption{Classification accuracies (\%) on \textbf{Digits} dataset.}
		\label{table:digit}
	\vspace{-5mm}
\end{table}

\section{Experiments}

\begin{table*}[!h]
    \begin{center}
    \scalebox{0.74}{  
         \begin{tabular}{lcccccccccccccc}
         \toprule
         Method & Source-free & Ar$\rightarrow$Cl & Ar$\rightarrow$Pr & Ar$\rightarrow$Rw & Cl$\rightarrow$Ar & Cl$\rightarrow$Pr & Cl$\rightarrow$Rw & Pr$\rightarrow$Ar & Pr$\rightarrow$Cl & Pr$\rightarrow$Rw & Rw$\rightarrow$Ar & Rw$\rightarrow$Cl & Rw$\rightarrow$Pr & Avg.\\
         \midrule
         ResNet-50 & \xmark & 34.9 & 50.0 & 58.0 & 37.4 & 41.9 & 46.2 & 38.5 & 31.2 & 60.4 & 53.9 & 41.2 & 59.9 & 46.1 \\
         MCD & \xmark & 48.9 & 68.3 & 74.6 & 61.3 & 67.6 & 68.8 & 57.0 & 47.1 & 75.1 & 69.1 & 52.2 & 79.6 & 64.1 \\
         CDAN & \xmark & 50.7 & 70.6 & 76.0 & 57.6 & 70.0 & 70.0 & 57.4 & 50.9 & 77.3 & 70.9 & 56.7 & 81.6 & 65.8 \\
         MDD & \xmark & 54.9 & 73.7 & 77.8 & 60.0 & 71.4 & 71.8 & 61.2 & 53.6 & 78.1 & 72.5 & 60.2 & 82.3 & 68.1 \\
         BNM & \xmark & 52.3 & 73.9 & 80.0 & 63.3 & 72.9 & 74.9 & 61.7 & 49.5 & 79.7 & 70.5 & 53.6 & 82.2 & 67.9 \\
         BDG & \xmark & 51.5 & 73.4 & 78.7 & 65.3 & 71.5 & 73.7 & 65.1 & 49.7 & 81.1 & 74.6 & 55.1 & 84.8 & 68.7 \\
         SRDC & \xmark & 52.3 & 76.3 & 81.0 & 69.5 & 76.2 & 78.0 & 68.7 & 53.8 & 81.7 & 76.3 & 57.1 & 85.0 & 71.3 \\
         \midrule
         PrDA & \cmark & 48.4 & 73.4 & 76.9 & 64.3 & 69.8 & 71.7 & 62.7 & 45.3 & 76.6 & 69.8 & 50.5 & 79.0 & 65.7 \\
        SHOT-IM & \cmark & 55.4 & 76.6 & 80.4 & 66.9 & 74.3 & 75.4 & 65.6 & 54.8 & 80.7 & 73.7 & 58.4 & 83.4 & 70.5 \\
         BAIT & \cmark & 57.4 & 77.5 & \textbf{82.4} & 68.0 & 77.2 & 75.1 & 67.1 & 55.5 & 81.9 & 73.9 & 59.5 & 84.2 & 71.6 \\
         CPGA & \cmark & \textbf{59.3} & \textbf{78.1} & 79.8 & 65.4 & 75.5 & 76.4 & 65.7 & \textbf{58.0} & 81.0 & 72.0 & \textbf{64.4} & 83.3 & 71.6 \\
        SHOT & \cmark & 57.1 & 78.1 & 81.5 & 68.0 & \textbf{78.2} & \textbf{78.1} &  67.4 & 54.9 & \textbf{82.2} & 73.3 & 58.8 & \textbf{84.3} & 71.8 \\ 
         \midrule
         \textbf{PS (ours)} & \cmark & 57.8 	&{77.3} 	& 81.2 	& \textbf{68.4} 	&{76.9} 	& \textbf{78.1} 	& \textbf{67.8} 	& 57.3 	&82.1 	& \textbf{75.2} 	& 59.1 	&{83.4} 	&\textbf{72.1} \\
         \bottomrule
         \end{tabular}}
         \vspace{-2mm}
         \caption{Classification accuracies (\%) on the \textbf{Office-Home} dataset (ResNet-50). 
         \label{tab:office-home}
    \vspace{-5mm}
    }
    \end{center}
\end{table*}

\begin{table*}[!hbt]
    \begin{center}
    \scalebox{0.85}{
         \begin{tabular}{lcccccccccccccc}
         \toprule
         Method & Source-free & plane & bicycle & bus & car & horse & knife & mcycl & person & plant & sktbrd & train & truck & Per-class\\
         \midrule
         ResNet-101 & \xmark & 55.1 & 53.3 & 61.9 & 59.1 & 80.6 & 17.9 & 79.7 & 31.2 & 81.0 & 26.5 & 73.5 & 8.5 & 52.4 \\
         CDAN & \xmark & 85.2 & 66.9 & 83.0 & 50.8 & 84.2 & 74.9 & 88.1 & 74.5 & 83.4 & 76.0 & 81.9 & 38.0 & 73.9 \\
         SAFN & \xmark & 93.6 & 61.3 & 84.1 & 70.6 & 94.1 & 79.0 & 91.8 & 79.6 & 89.9 & 55.6 & 89.0 & 24.4 & 76.1 \\
         SWD & \xmark & 90.8 & 82.5 & 81.7 & 70.5 & 91.7 & 69.5 & 86.3 & 77.5 & 87.4 & 63.6 & 85.6 & 29.2 & 76.4 \\
         TPN & \xmark & 93.7 & 85.1 & 69.2 & 81.6 & 93.5 & 61.9 & 89.3 & 81.4 & 93.5 & 81.6 & 84.5 & 49.9 & 80.4 \\
         PAL & \xmark & 90.9 & 50.5 & 72.3 & 82.7 & 88.3 & 88.3 & 90.3 & 79.8 & 89.7 & 79.2 & 88.1 & 39.4 & 78.3 \\    
         MCC & \xmark & 88.7 & 80.3 & 80.5 & 71.5 & 90.1 & 93.2 & 85.0 & 71.6 & 89.4 & 73.8 & 85.0 & 36.9 & 78.8 \\
         CoSCA & \xmark & 95.7 & 87.4 & 85.7 & 73.5 & 95.3 & 72.8 & 91.5 & 84.8 & 94.6 & 87.9 & 87.9 & 36.8 & 82.9 \\
         \midrule
         PrDA & \cmark & 86.9 & 81.7 & \textbf{84.6} & 63.9 & 93.1 & 91.4 & 86.6 & 71.9 & 84.5 & 58.2 & 74.5 & 42.7 & 76.7\\
        SHOT-IM & \cmark & 93.7 & 86.4 & 78.7 & 50.7 & 91.0 & 93.5 & 79.0 & 78.3 & 89.2 & 85.4 & 87.9 & 51.1 & 80.4 \\
         MA & \cmark & 94.8 & 73.4 & 68.8 & \textbf{74.8} & 93.1 & 95.4 & \textbf{88.6} & \textbf{84.7} & 89.1 & 84.7 & 83.5 & 48.1 & 81.6\\
         BAIT & \cmark & 93.7 & 83.2 & 84.5 & 65.0 & 92.9 & 95.4 & 88.1 & 80.8 & 90.0 & 89.0 & 84.0 & 45.3 & 82.7 \\
        SHOT   & \cmark  & 94.3 & \textbf{88.5} & 80.1 & 57.3 & 93.1 & 94.9 & 80.7 & 80.3 & 91.5 & 89.1 & \textbf{86.3} & 58.2 & 82.9 \\
        \midrule
        \textbf{PS (ours)} & \cmark & \textbf{95.3} & 86.2 & {82.3} & 61.6 & \textbf{93.3} & \textbf{95.7} & 86.7 &{80.4} & \textbf{91.6} & \textbf{90.9} & 86.0  &  \textbf{59.5} & \textbf{84.1} \\
        \bottomrule
        \end{tabular}
    }
    \vspace{-1mm}
    \caption{ Classification accuracies (\%) on the large-scale \textbf{VisDA} dataset (ResNet-101).}
    \label{tab:visda}
    \end{center}
    \vspace{-6mm}
\end{table*}

\subsection{Datasets}\label{des:dataset}

We conduct experiments on three benchmark datasets:
(1) \textbf{Digits} is a standard dataset that focuses on digit recognition. We follow the protocol of \cite{Hoffman2018CyCADACA} and choose three subsets: SVHN (S), MNIST (M), and USPS (U). 
(2) \textbf{Office-Home}~\cite{Venkateswara2017DeepHN} is a widely used dataset, which contains four domains: Artistic images (Ar), Clip Art (Cl), Product images (Pr) and Real-world images (Rw). Each  domain has 65 categories.
(3) \textbf{VisDA}~\cite{Peng2017VisDATV} is a challenging large-scale dataset that concentrates on the synthesis-to-real object recognition task. The source domain contains 152k synthetic images 
while the target domain has 55k real object images with 12 classes.

\subsection{Baselines}

We compare PS with three types of baselines:
(1) \textbf{Source-only}: ResNet~\cite{He2016DeepRL} or LeNet \cite{LeCun1998GradientbasedLA};
(2) \textbf{ UDA methods}:
ADDA \cite{Tzeng2017AdversarialDD},
 ADR \cite{Saito2018AdversarialDR},
 MCD~\cite{saito2018maximum}, CDAN~\cite{long2018conditional},
 CyDADA \cite{Hoffman2018CyCADACA},
  SAFN~\cite{xu2019larger}, SWD~\cite{lee2019sliced}, TPN~\cite{pan2019transferrable},
 CAT \cite{Deng2019ClusterAW},
 MDD~\cite{zhang2019bridging}, SWD~\cite{Lee2019SlicedWD}, BDG~\cite{yang2020bi}, PAL~\cite{hu2020panda}, MCC~\cite{jin2020minimum},
 BNM~\cite{Cui2020TowardsDA},
 CoSCA~\cite{Dai_2020_ACCV} and SRDC~\cite{tang2020unsupervised}; (3) \textbf{ SFDA methods}: 
 PrDA~\cite{kim2020progressive},
 SHOT~\cite{liang2020shot},  MA~\cite{Li2020ModelAU},
 BAIT~\cite{Yang2020UnsupervisedDA} and
 CPGA \cite{Qiu2021SourcefreeDA}.

\subsection{Implementation Details}

Our method is implemented based on PyTorch. 
For a fair comparison, we report the results of all baselines in the corresponding papers.
For the network architecture, a ResNet~\cite{He2016DeepRL} pre-trained on ImageNet is adopted as the  feature extractor of all methods for the object recognition task. For the digit recognition task, we use the same architectures with CDAN \cite{long2018conditional}, namely, the classical LeNet-5 \cite{LeCun1998GradientbasedLA} network is utilized for USPS$\leftrightarrow$ MNIST and a variant of LeNet is utilized for SVHN$\rightarrow$ MNIST. Following~\cite{liang2020shot}, we replace the original fully connected (FC) layer with a task-specific FC layer followed by a weight normalization layer. 
We train the model using SGD optimizer with momentum 0.9 and weight decay $5e^{-4}$. 
The learning rate of the domain discriminator and the classifier is 10 times that of the feature extractor, where the former is set to be $1e^{-4}$ for Digits and $1e^{-5}$ for Office-Home and VisDA.
Besides, the number of epoch is set to be 200, 800, and 60 for Digits, Office-Home, and VisDA, respectively.
For hyper-parameters, we set $\alpha$ to be 0.1 for all datasets and batchsize $B$ to be 500, 658, and 128 for Digits, Office-Home, and VisDA,  respectively. We set $\lambda_g=0.5$ and $\lambda_{c}=1$ for all datasets except $\lambda_{c}=0.3$ for Office-Home.
%

\subsection{Results}
For \textbf{Digits} recognition,  as shown in Table \ref{table:digit}, PS obtains the best mean accuracies for each task and outperforms prior work in terms of the average accuracy. The proposed method performs better than SHOT and SHOT-IM, as PS could take the distribution discrepancy into consideration. It is noticed that both PS and SHOT achieve better results than the supervised learning for task M$\rightarrow$U, which is because that the large source domain MNIST contains more useful information suitable for the small target domain USPS. For \textbf{Office-Home} dataset, as shown in Table \ref{tab:office-home}, the proposed PS achieves the best performance compared with other SFDA methods w.r.t. the average accuracy over four transfer tasks. Moreover, our method shows its superiority in the task of Cl$\rightarrow$Ar and Rw$ \rightarrow$Ar and comparable results on the other tasks. Moreover, from Table \ref{tab:visda}, PS outperforms all the state-of-the-art methods on the more challenging large-scale dataset \textbf{VisDA}. Specifically, PS gets the best accuracy in the six categories and obtains comparable results in others.
Note that for these three datasets, even compared with the state-of-the-art methods using source data (e.g.,SWD, SRDC, CoSCA), our PS is able to obtain a competitive result as well, which shows that PS could make a better adaptation across domains.

\subsection{Insight Analysis}

\subsubsection{Ablation of selection  criterion of pseudo-source samples}
To study the selection criterion of pseudo-source samples, we compare the model with two criterion, i.e., entropy criterion (our method) and random selection (with the same  proportion in each class). As shown in Table \ref{ab:com}, PS with entropy criterion (line 3) outperforms random selection (line 1) by 0.5\% in VisDA dataset. We also find that random selection achieves better results than some SFDA method, as such split strategy can be regraded as reducing the intra-domain discrepancy within the target domain \cite{Pan2020UnsupervisedIA}.

\begin{table}[t]
\begin{center}
\scalebox{0.91}{
\begin{tabular}{lcc}
		\toprule
		Method& Digits & VisDA\\
		\midrule
		{Random selection with mixup}     & 98.4 & {83.6}\\
		{Entropy criterion without  mixup}  & 98.4  & 83.4 \\
		\midrule
		\textbf{Entropy criterion with mixup (ours)}                  & \textbf{98.6} & \textbf{84.1}\\
		\bottomrule
		\end{tabular}
		}
		 \vspace{-1mm}
		\caption{Ablation study of generation and augmentation on Digits and VisDA dataset. }
		\label{ab:com}
    \end{center}
    \vspace{-7mm}
\end{table}

\subsubsection{Ablation of pseudo-source domain augmentation}
To explore the impact of data augmentation, we conduct experiments with and without mixup. From Table \ref{ab:com}, compared with other SFDA method, both settings achieve better performance. In other words, even without mixup (line 2), our method could also get comparable results. With mixup (line 3), the model gets a better performance.

\subsubsection{Ablation of loss functions}

To investigate the losses in distribution alignment, the quantitative results of the model optimized by different losses are shown in Table \ref{tab:Ablation}. When only using classification loss $\mathcal{L}_{cls}$, the model gets better performance than source-only model as the self-training based strategy could perform implicit alignment to the source model. Besides, our model is able to  further improve the performance when introducing the losses $\mathcal{L}_{div}$ and $\mathcal{L}_{con}$. Such a result verifies the benefit of encouraging diverse outputs and retaining knowledge of source domain. Finally, a combination of all losses can achieve the best results.

\subsubsection{Distribution distance} 

\begin{figure}[t]
	\centering
	\includegraphics[width=0.29\textwidth]{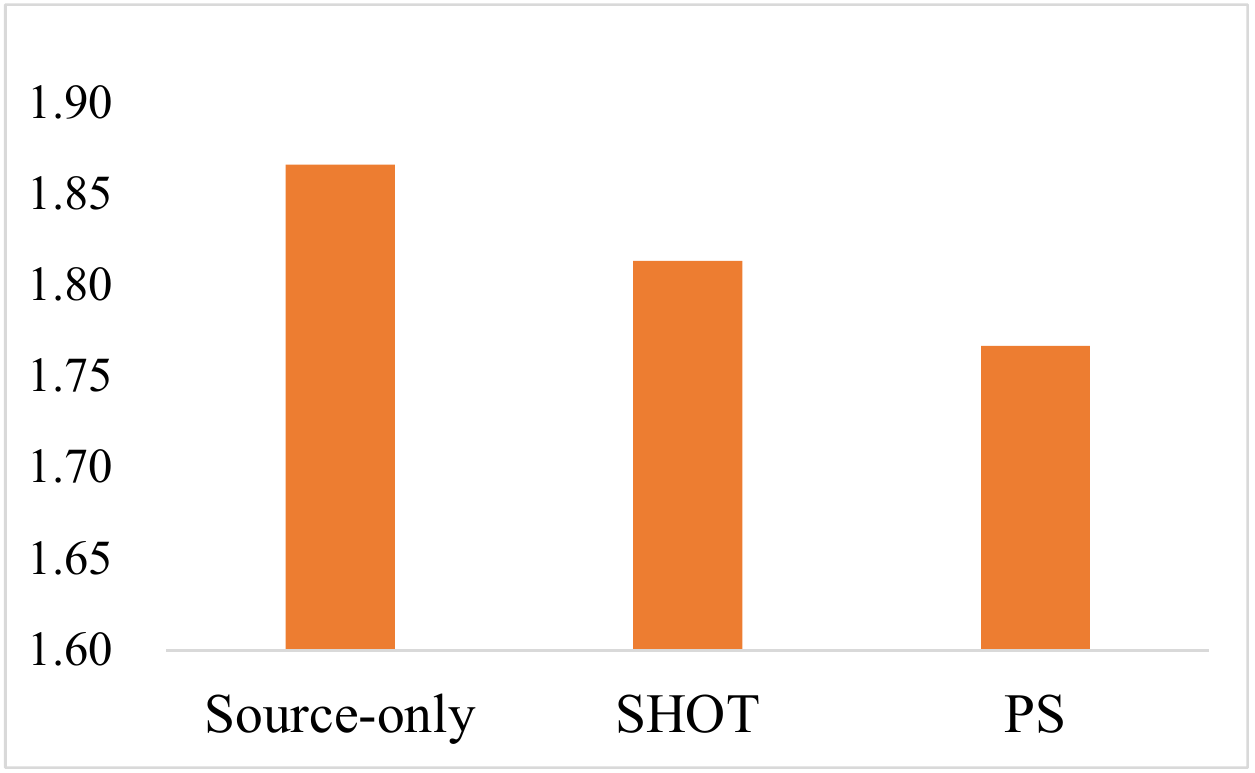}
	\vspace{-3mm}
	\caption{$\mathcal{A}$-distance.}
	\label{fig:a_dis}
	\vspace{-5mm}
\end{figure}

The $\mathcal{A}$-distance is a measure of domain discrepancy \cite{BenDavid2009ATO}, defined as $d_A = 2(1-2\epsilon)$,  where $\epsilon$ is the error rate of a domain classifier trained to discriminate source domain and target domain. To show the true distribution distance across domains, we compute the $\mathcal{A}$-distance using both true source samples and target samples. Note that the source samples are only used for computing $\mathcal{A}$-distance, not used for training. Results on tasks M$\rightarrow$U are shown in Figure \ref{fig:a_dis}. SHOT implicitly aligns representations of the target domain to the source model, thus getting a smaller distance than source-only method. PS could get smaller $\mathcal{A}$-distance than SHOT by further reducing the domain shift explicitly.

\begin{table}[b]
\begin{center}
\scalebox{0.9}{
\begin{tabular}{lcc}
		\toprule
		Method & Digits & VisDA \\
		\midrule
		Source-only &79.3 &46.6 \\
		$\mathcal{L}_{cls}$ & 98.3 & 77.6 \\ 
		$\mathcal{L}_{cls}$ +  $\mathcal{L}_{div}$ & 98.3 & 78.4 \\
		$\mathcal{L}_{cls}$ +  $\mathcal{L}_{div}$ + $\mathcal{L}_{cons}$  & 98.4 & 83.0 \\
		Ours (all)                   & \textbf{98.6}  & \textbf{84.1} \\
		\bottomrule
		\end{tabular}
		}
\end{center}
\vspace{-2mm}
	\caption{Ablation of losses on Digits and VisDA dataset.} 
	\label{tab:Ablation}
	\vspace{-3mm}
\end{table}

\subsubsection{Parameter sensitivity}
\begin{figure}[t]
	\centering
	\subfigure[Sensitivity of batchsize $B$]{
	\includegraphics[width = 0.23\textwidth]{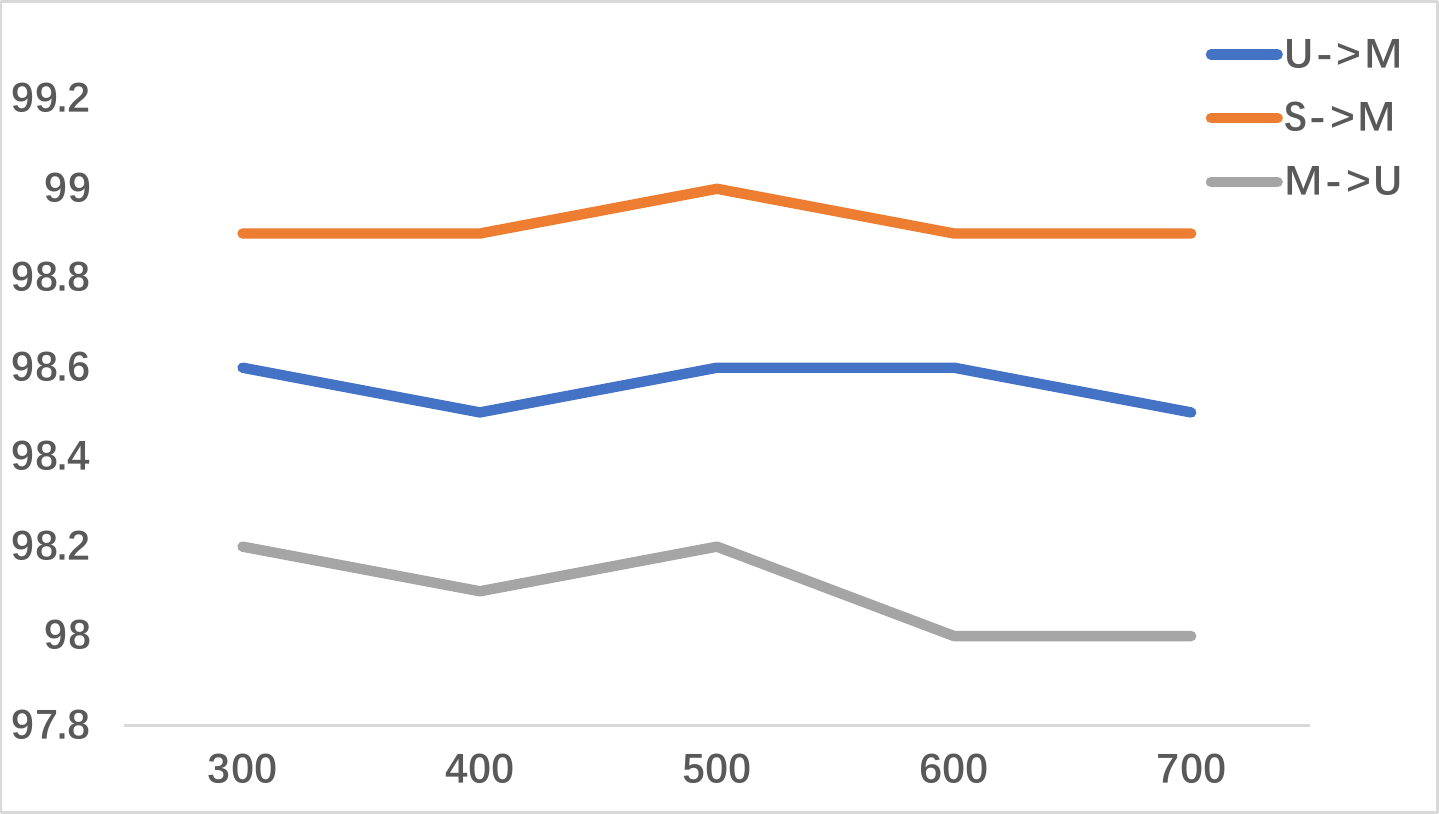}
	}
	\subfigure[Sensitivity of  proportion $\alpha$]{
	\includegraphics[width = 0.205\textwidth]{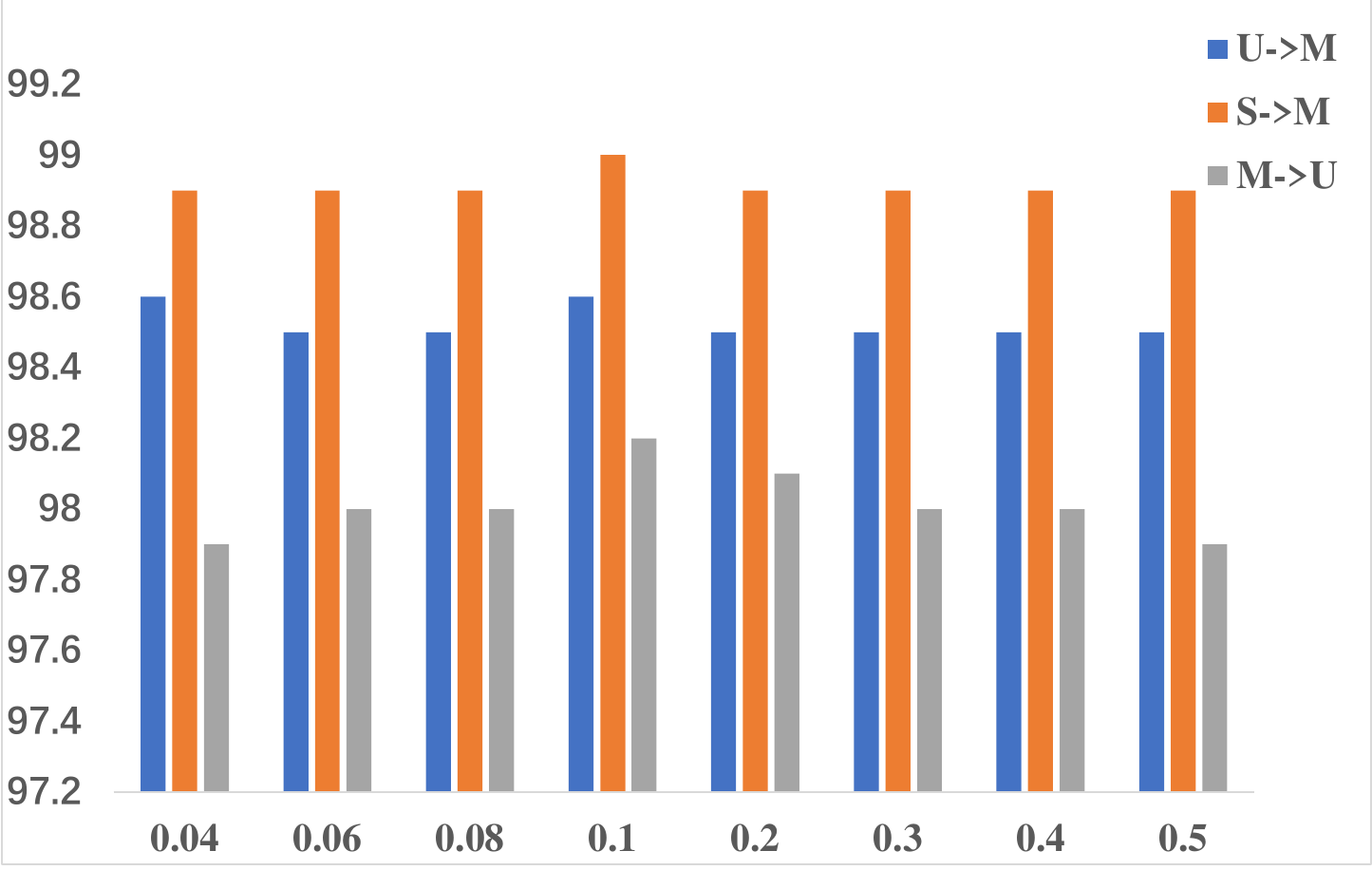}
    }
    \vspace{-3mm}
    \caption{Results of parameter sensitivity.}
    \label{fig_sen}
    \vspace{-5mm}
\end{figure}

In this subsection, we evaluate the sensitivity of two hyper-parameters batchsize $B$ and  proportion $\alpha$ on Digits. As shown in Figure \ref{fig_sen}(a), the results demonstrate that our method is non-sensitive to the batchsize. Although PS generates and augments the pseudo-source domain within a minibatch instead of the whole domain, it could achieve stable results because many random minibatches can be seen as an approximation of the whole domain. The results in Figure \ref{fig_sen}(b) show that with the increasing of $\alpha$, the performance grows firstly and then drops.  The reason is that a small  proportion would lead to a small number of pseudo-source samples and a large  proportion would select some samples not similar to the source domain into the pseudo-source domain. We experimentally find $\alpha=0.1$ works well.

\subsubsection{Embedding visualization 
}
\begin{figure}[h]
	\centering
	\includegraphics[width=0.48\textwidth]{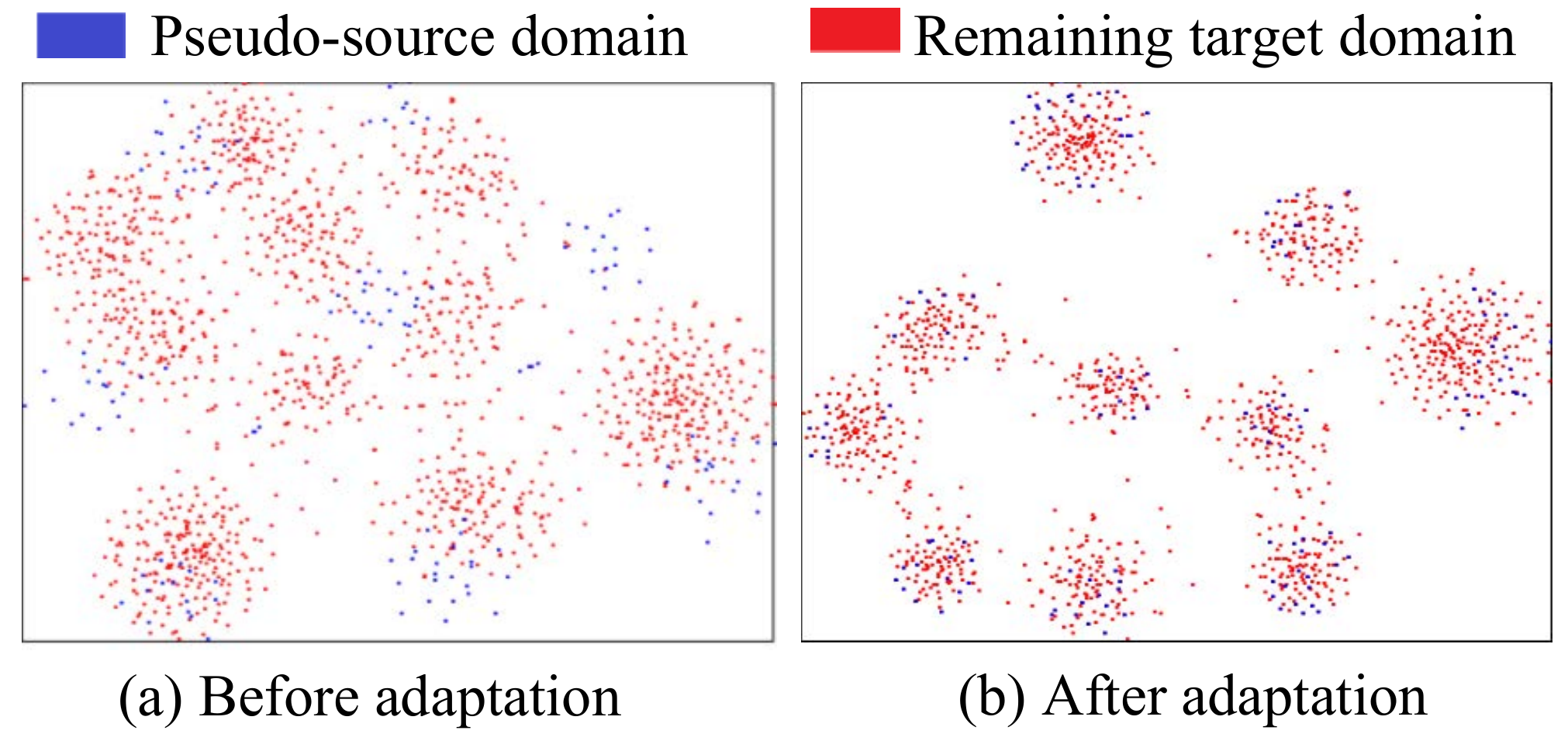}
	\vspace{-3mm}
	\caption{t-SNE visualization for features from the pseudo-source domain and the remaining target domain.}
	\label{fig:tsne}
	\vspace{-2mm}
\end{figure}

Figure \ref{fig:tsne}(a) and (b) show the t-SNE visualization of the features from the pseudo-source domain and the remaining target domain for task M$\rightarrow$U (10 classes) before and after alignment, respectively. 
 Before alignment, there exists large distribution shift between the pseudo-source domain and the remaining target domain.
While after alignment the distribution shift is reduced and the features of target samples have become discriminative,  thus, the samples can be easily classified by the classifier.

\section{Conclusions}
In this paper, we focus on source-free domain adaptation, where only a trained source model and unlabeled target samples are given. Considering that previous SFDA methods do not  explicitly  reduce  the  distribution  shift  across  domains, we propose a novel method to reduce the shift explicitly. To be specific, we find that some target samples can be used to represent the source domain and the new domain is denoted as pseudo-source domain. The proposed method firstly generates and augments the pseudo-source domain and then employs distribution alignment between the pseudo-source domain and the remaining target domain. Thus, the domain shift can be reduced. Extensive experiments on three datasets show that our method outperforms the state-of-the-art methods.

\newpage
\bibliography{aaai2021}

\end{document}